\documentclass{vgtc}                          

\ifpdf
  \pdfoutput=1\relax                   
  \usepackage{graphicx}                
  \DeclareGraphicsExtensions{.pdf,.png,.jpg,.jpeg} 
\else
  \usepackage{graphicx}                
  \DeclareGraphicsExtensions{.eps}     
\fi%

\graphicspath{{figures/}{pictures/}{images/}{./}} 
\usepackage{amssymb}

\usepackage{fancyvrb}
\usepackage{xcolor}

\DefineVerbatimEnvironment{MetaVerbatim}
  {Verbatim}
  {formatcom=\color{codecommentcolor}}
\colorlet{codecommentcolor}{green!60!red}

\usepackage{microtype}                 
\PassOptionsToPackage{warn}{textcomp}  
\usepackage{textcomp}                  
\usepackage{mathptmx}                  
\usepackage{times}                     
\usepackage{cite}                      
\usepackage{tabu}                      
\usepackage{booktabs}                  
\usepackage{mathptmx}                  
\usepackage{algorithm}
\usepackage{algpseudocode}
\usepackage{multirow}

\definecolor{object}{rgb}{1, 0.49, 0.0}
\definecolor{context}{rgb}{0.333, 0.447, 0.753}

\onlineid{1026}

\vgtccategory{Research}

\vgtcinsertpkg

\title{Visual Attention Exploration in Vision-Based Mamba Models}

\author{Junpeng Wang\thanks{e-mail: junpenwa@visa.com}\\ %
        \scriptsize Visa Research %
\and Chin-Chia Michael Yeh\thanks{e-mail: miyeh@visa.com}\\ %
     \scriptsize Visa Research %
\and Uday Singh Saini\thanks{e-mail: udasaini@visa.com}\\ %
     \parbox{1.4in}{\scriptsize \centering Visa Research}
\and Mahashweta Das\thanks{e-mail: mahdas@visa.com}\\ %
     \parbox{1.4in}{\scriptsize \centering Visa Research}
     }

\teaser{
  \centering
  \vspace{-0.17in}
  \includegraphics[width=\linewidth]{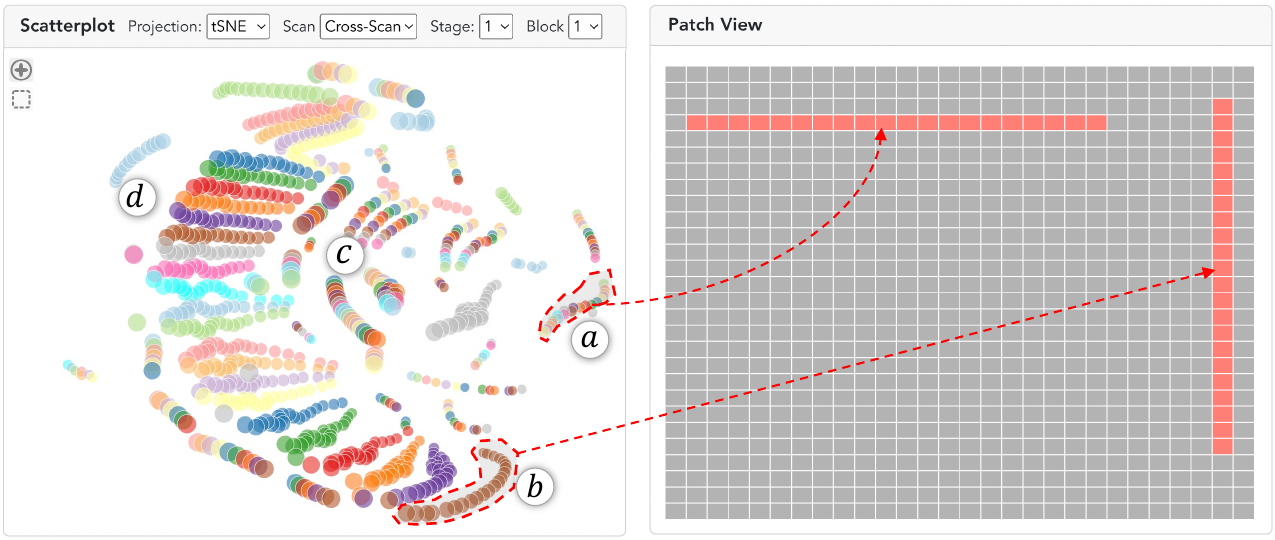}
  \vspace{-0.3in}
  \caption{Our visual exploration tool contains two visualization components. The \textit{Scatterplot} view on the left shows the dimensionality reduction results. The \textit{Patch} view on the right shows the patch layout and highlights the patches of interest.}
  \label{fig:system}
}

\abstract{%
State space models (SSMs) have emerged as an efficient alternative to transformer-based models, offering linear complexity that scales better than transformers. One of the latest advances in SSMs, Mamba, introduces a \textit{selective scan} mechanism that assigns trainable weights to input tokens, effectively mimicking the attention mechanism. Mamba has also been successfully extended to the vision domain by decomposing 2D images into smaller patches and arranging them as 1D sequences. However, it remains unclear how these patches interact with (or attend to) each other in relation to their original 2D spatial location. Additionally, the order used to arrange the patches into a sequence also significantly impacts their attention distribution. To better understand the attention between patches and explore the attention patterns, we introduce a visual analytics tool specifically designed for vision-based Mamba models. This tool enables a deeper understanding of how attention is distributed across patches in different Mamba blocks and how it evolves throughout a Mamba model. Using the tool, we also investigate the impact of different patch-ordering strategies on the learned attention, offering further insights into the model's behavior.
}

\begin{document}


\firstsection{Introduction}

\maketitle

State space models (SSMs) use state variables to mathematically describe the state of a dynamic system. 
They have a long history of modeling time series problems, where the state variables are time-dependent. Recent advances~\cite{gu2021efficiently, smith2022simplified, gu2023mamba} have shown that SSMs achieve performance on par with state-of-the-art transformer models~\cite{vaswani2017attention, dosovitskiy2020image}. Additionally, their linear time complexity allows them to outperform transformers in latency-critical applications.

Mamba~\cite{gu2023mamba}, also known as S6, is a cutting-edge SSM that enhances its predecessor S4~\cite{gu2021efficiently} with a selective scan mechanism. This mechanism enables the model to assign trainable weights to input tokens, allowing it to filter out less relevant information and emphasize more relevant details. These trainable weights are similar to the attention mechanism in transformers, which determines how much focus a token should place on other tokens. Since its release in late 2023, vision-based applications of Mamba have been rapidly developed. 
Similar to how transformers are adapted for vision tasks~\cite{dosovitskiy2020image}, vision-based Mamba models also decompose images into smaller patches and arrange them into sequences as inputs.
\begin{figure}[tbh]
 \centering 
\includegraphics[width=\columnwidth]{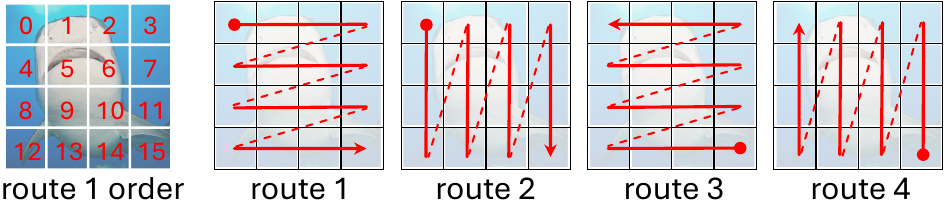}
 \caption{The four-way cross-scan in VMamba~\cite{liu2024vmamba}.}
 \label{fig:patch}
\end{figure}

In vision transformers~\cite{dosovitskiy2020image}, all image patches are fed into the models simultaneously, with positional encoding used to differentiate patches at different positions. As a result, the order of patches has a relatively minor impact. However, in vision-based Mamba models, the patches are fed into the models sequentially one after another (similar to RNNs). This means that the order of the patches is crucial, as each patch can only collect information from its preceding ones. To ensure that each patch has access to all other patches, both forward and backward scans are often performed. For example, vision-Mamba (Vim)\cite{zhu2024vision} uses routes 1 and 3 in Fig.~\ref{fig:patch} to allow each patch to ``see'' both its preceding and succeeding patches. VMamba~\cite{liu2024vmamba} uses all four routes in Fig.~\ref{fig:patch} to preserve spatial locality between patches within the same row and column.

A vision-based Mamba model consists of a hierarchy of Mamba blocks, each of which learns the attention between image patches. However, it remains unclear (1) if the attention pattern is fixed within a block, (2) how attention patterns evolve across blocks, and (3) how patch-ordering strategies impact the attention pattern. This paper seeks to address these questions using a visual analytics approach. Specifically, we extract the attention learned at each stage of the vision-based Mamba model and apply dimensionality reduction techniques to reveal attention patterns across stages. We also profile the attention patterns at each patch to disclose how attention is spatially distributed relative to the patch's position. Lastly, we introduce different patch-ordering strategies and compare their resulting attention patterns. In summary, our contributions are twofold: 
\begin{enumerate} 
\item We design and develop a visual analytics tool to explore and summarize attention patterns in vision-based Mamba models. 
\item We introduce multiple patch-ordering strategies and investigate their impact on patch attention in vision-based Mamba models. 
\end{enumerate}

\section{Background and Related Work}
Our work focuses on interpreting the attention mechanisms in vision-based Mamba models. Here, we briefly review the history of SSMs and summarize early work on attention interpretation.

\subsection{State Space Models (SSMs)}
State space models (SSMs) are sequence modeling techniques that date back to the 1960s. They gained significant attention after Gu et al.~\cite{gu2021efficiently} integrated the HiPPO matrix into them. The resulting model, known as S4, demonstrated substantial improvements in efficiency for modeling long sequences. However, the performance of S4 is not comparable to that of transformers, as it cannot effectively distinguish different input tokens. To overcome this, Gu and Dao~\cite{gu2023mamba} introduced the selective scan mechanism into S4, giving rise to S6, also known as Mamba. The selective scan allows Mamba to assign different weights to input tokens, enabling the model to learn the relative importance of each token. This allows Mamba to focus more on important tokens and selectively ignore less relevant ones. The learned weights are similar to the attention weights in transformers. Mamba demonstrates performance on par with transformers across a wide range of applications. More importantly, its linear complexity makes it a superior choice for many latency-critical applications, where the quadratic complexity of transformers would be prohibitive.

\begin{figure}[tbh]
 \centering 
\includegraphics[width=\columnwidth]{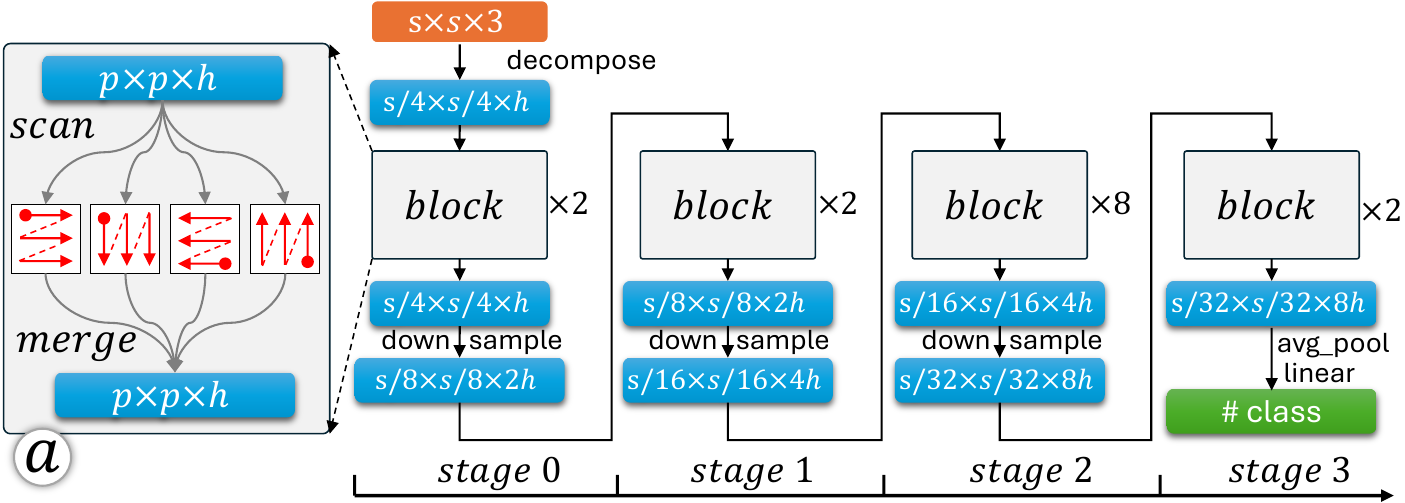}
 \caption{The architecture of VMamba for image classification~\cite{liu2024vmamba}.}
 \label{fig:mamba}
\end{figure}
In the vision domain, vision-Mamba (Vim)\cite{zhu2024vision} and VMamba\cite{liu2024vmamba} are two seminal works that extend Mamba from NLP to vision tasks. Fig.~\ref{fig:mamba} illustrates the architecture of VMamba for image classification. Given an RGB input image of size $s {\times} s$, the model first decomposes it into $p {\times} p$ smaller patches. Each patch is considered as a token, and all patches are arranged into a sequence of length $p {\times} p$. This sequence is then treated similarly to textual data and fed into Mamba for learning. 
Specifically, the VMamba block in Fig.~\ref{fig:mamba}a employs four different \texttt{scan} orders (Fig.~\ref{fig:patch}) to process the patches: (1) left$\rightarrow$right then top$\rightarrow$down, (2) top$\rightarrow$down then left$\rightarrow$right, (3) right$\rightarrow$left then bottom$\rightarrow$up, and (4) bottom$\rightarrow$up then right$\rightarrow$left. Each scan order introduces a unique dependency between patches, and consequently, the learned attention between patches also varies. The \texttt{merge} operation aggregates the attention learned for each patch across the four different scan orders. The output of this process is a $p^2 {\times} p^2$ attention matrix, where each element at position ($i$, $j$) represents the attention strength between patch $i$ and patch $j$. Stacking multiple \texttt{scan-merge} blocks forms a stage, and Fig.~\ref{fig:mamba} shows four such stages, each containing 2, 2, 8, and 2 blocks, respectively. Between stages, the latent representations are downsampled to distill essential features. Finally, a fully-connected layer transforms the latent representation into a vector, the length of which corresponds to the number of possible classes. Each element in this vector determines the probability for the corresponding class. Vim~\cite{zhu2024vision} follows a similar process, but uses only a two-way scan (routes 1 and 3 in Fig.~\ref{fig:patch}). Therefore, we focus on VMamba in this work.

\subsection{Attention Analysis and Visualization}
Interpreting and diagnosing machine learning models is an important topic in the visualization literature~\cite{wang2024visual, hohman2018visual, Liu2025}, and there are multiple works focusing on interpreting the attention mechanisms of these models. For example, Abnar and Zuidema~\cite{abnar2020quantifying}, Park et al.~\cite{park2019sanvis}, and Li et al.~\cite{li2021t3} all used heatmaps to externalize the attention strengths in transformer-based NLP models. Vig~\cite{vig2019multiscale} employed multiple parallel coordinate plots to visualize the attention patterns in BERT and GPT-2 models. DeRose et al.~\cite{derose2020attention} introduced a radial layout for visualizing attention in BERT, layer by layer, which also facilitates comparisons between the attentions of two models.
For vision transformers, Li et al.~\cite{li2023does} used dimensionality reduction and scatterplots to summarize attention patterns within self-attention heads and across attention layers. Yeh et al.~\cite{yeh2023attentionviz} employed a similar approach to visualize the joint $q/k$ embedding space, providing a global view of attention patterns across a transformer model.

To the best of our knowledge, no studies have yet focused on analyzing the attention patterns in Mamba models. In particular, we are interested in whether regular attention patterns emerge within a Mamba block and whether hierarchical attention patterns develop across different stages of a Mamba model. To explore these questions, we have developed a visual analytics tool specifically designed to analyze attention patterns in vision-based Mamba models.

\section{Problem and Methodology}
In this study, we aim to investigate the attention patterns in Mamba, focusing on two key aspects based on insights from existing literature and the interests of domain experts working with Mamba:
\begin{itemize} 
\item \textit{Inter-Block Attention Pattern}: Are the attention patterns consistent across different Mamba blocks within the same stage?  
\item \textit{Intra-Block Attention Pattern}: How is the attention spatially distributed across image patches within a single Mamba block? 
\end{itemize}

We provide solutions to answer these two questions in Sec.~\ref{sec:stage} and Sec.~\ref{sec:block}, respectively. For our study, we focus on the VMamba model~\cite{liu2024vmamba}, whose architecture is illustrated in Fig.~\ref{fig:mamba}. The model is trained for 400 epochs on the ImageNet dataset (1,281,167 training images). After training, we use 1000 test images to investigate the attention patterns. The input images are of size $224{\times}224{\times}3$ (i.e., $s{=}224$ in Fig.~\ref{fig:mamba}) and the number of patches at the four stages is $56{\times}56$, $28{\times}28$, $14{\times}14$, and $7{\times}7$, respectively.

\subsection{Inter-Block Attention Pattern}
\label{sec:stage}

The attention matrices for different Mamba blocks at the same stage have the same size. For example, in Fig.~\ref{fig:mamba}, the attention matrices from the two blocks at stage 0 are of the same size, which is $56^2{\times}56^2$ when $s {=} 224$. This allows us to compare the matrices and assess how the attention patterns differ across blocks at the same stage. However, attention matrices from blocks across different stages have varying sizes, and thus, these blocks are not directly comparable.

When an image is fed through a Mamba block, it produces an attention matrix of size \(p^2 {\times} p^2\), where the element at position \((i, j)\) represents the attention strength from patch \(i\) to patch \(j\). 
For example, at stage 0 in Fig.~\ref{fig:mamba}, the attention matrix will be of shape $56^2{\times}56^2$.
If the focused stage contains $m$ blocks, we will obtain $m$ such \(p^2 {\times} p^2\) matrices.
By feeding all \(n{=}1000\) test images through the VMamba model and collecting their attention matrices from the $m$ blocks at the focused stage, we compile the attention data into a large matrix of shape $(m {\times} n) {\times} (p^2{\times} p^2)$. This process is outlined in lines 1 to 7 of the pseudo-code in Algorithm~\ref{alg:stage}.

We then apply dimensionality reduction (DR) techniques to map the $m {\times} n$ points, originally in the $p^2{\times} p^2$ dimensional space, to a 2D space for visualization as a scatterplot. If the $n$ points from different blocks form isolated clusters, this indicates that the attention patterns between the blocks are significantly different. The details of the visualization and interactions are described later in Sec.~\ref{sec:scatterplot}.

\begin{algorithm}[t]
\caption{Generating cluster pattern across blocks of a stage.}
\label{alg:stage}
\begin{algorithmic}[1]
\Require $n, m$ \Comment{number of images and blocks}
\Require $stage$ \Comment{the focused stage}
\Require $image$ \Comment{the array of images}
\Require $VMamba, DR$ \Comment{the VMamba model and DR algorithm}
\State $attentions = []$
\For{$i \gets 1$ to $n$}
    \For{$j \gets 1$ to $m$}
        \State $attention = VMamba(image[i], stage, j)$
        \State $attentions.append(attention)$
    \EndFor
\EndFor
\State $points = DR(attentions)$ \Comment{$points \in \mathbb{R}^{m{\times}n{\times}2}$}
\end{algorithmic}
\end{algorithm}

\subsection{Intra-Block Attention Pattern}
\label{sec:block}
According to the formulation in Sec.~\ref{sec:stage}, the attention matrix for a single block and a single image has the shape $p^2 {\times} p^2$ (line 4 of Algorithm~\ref{alg:stage}). The $i$-th row of this matrix represents the attention strength from patch $i$ to all $p^2$ patches. To identify patches with similar attention patterns, we can apply dimensionality reduction (DR) techniques to reduce the $p^2 {\times} p^2$ matrix to a $p^2 {\times} 2$ matrix, which can then be analyzed through a scatterplot visualization to assess the similarity between patches.

However, the approach described above can be significantly influenced by the image content, making it difficult to extract common, content-agnostic patterns. To enhance the identification of common attention patterns, we generate the $p^2 {\times} p^2$ attention matrix for all \(n\) images and aggregate the resulting $n$ matrices into a single $p^2 {\times} p^2$ matrix for pattern augmentation. The details of this process are outlined in Algorithm~\ref{alg:block} (lines 1 to 6).

We employ a scatterplot to visualize the \(p^2\) 2D points. The color and size of each point correspond to the column and row of the respective patch. By examining the clustering patterns of these points, we can identify patches with similar attention patterns and relate these similarities to their spatial position in the image space (detailed in Sec.~\ref{sec:scatterplot}).

\begin{algorithm}[tbh]
\caption{Generating cluster pattern within a block.}
\label{alg:block}
\begin{algorithmic}[1]
\Require $n$ \Comment{number of images}
\Require $stage, block$ \Comment{the focused stage and block}
\Require $image$ \Comment{the array of images}
\Require $VMamba, DR$ \Comment{the VMamba model and DR algorithm}
\State $attentions = []$
\For{$i \gets 1$ to $n$}
        \State $attention = VMamba(image[i], stage, block)$
        \State $attentions.append(attention)$
\EndFor
\State $avg\_attn = np.mean(np.array(attentions), axis=0)$
\State $points = DR(avg\_attn)$ \Comment{$points \in \mathbb{R}^{p^2{\times}2}$}
\end{algorithmic}
\end{algorithm}

\section{Visual Analytics System}
We have developed a visual analytics tool to effectively visualize the outputs of Algorithm~\ref{alg:stage} and Algorithm~\ref{alg:block}. The tool features two distinct visualization views: the \textit{Scatterplot} view (Fig.~\ref{fig:system}, left) and the \textit{Patch} view (Fig.~\ref{fig:system}, right).

\subsection{The \textit{Scatterplot} View}
\label{sec:scatterplot}
The \textit{Scatterplot} view provides two visualization modes for exploring the inter-block and intra-block attention patterns.

\begin{figure}[tbh]
 \centering 
\includegraphics[width=\columnwidth]{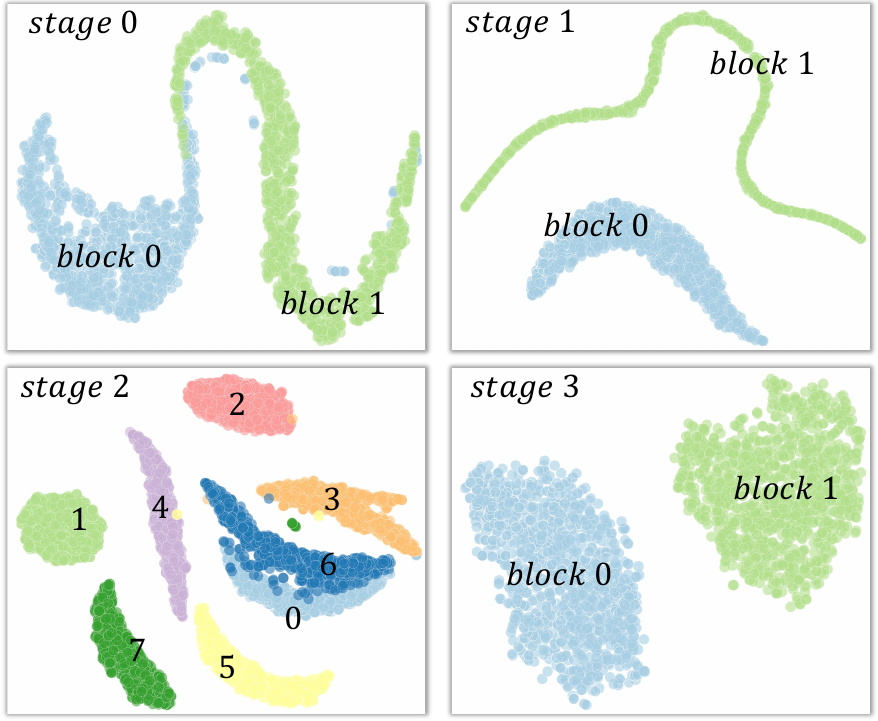}
 \caption{Attention pattern similarity between blocks from the four stages of the VMamba model in Fig.~\ref{fig:mamba}.}
 \label{fig:stage}
\end{figure}

\paragraph{Mode 1:} Users can select a stage and a block from the header of the \textit{Scatterplot} view (Fig.~\ref{fig:system}, left). If the block ID is not specified, the \textit{Scatterplot} will display $m {\times} n$ points based on the output of Algorithm~\ref{alg:stage}, where $m$ represents the number of blocks in the selected stage, and $n {=} 1000$ is the number of images. Each point corresponds to a $p^2 {\times} p^2$ attention matrix for a given image from a specific block.
For example, in the VMamba architecture depicted in Fig.~\ref{fig:mamba}, there are 4 stages, with 2, 2, 8, and 2 blocks per stage, respectively. The dimensionality reduction results for the four stages are shown in Fig.~\ref{fig:stage}, where each figure represents the output of a single stage. The color in each figure corresponds to the block ID, and distinct clusters of points are clearly visible, with different colors separating the blocks. This indicates that blocks within the same stage exhibit noticeably different attention patterns.

\paragraph{Mode 2:} When a block ID is specified, the \textit{Scatterplot} will show $p^2$ points based on the output of Algorithm~\ref{alg:block}. In this case, each point represents the averaged attention across $n{=}1000$ test images for a specific patch position. The color and size of each point correspond to the column and row of the corresponding patch, respectively. 
As illustrated by the clustering pattern in Fig.~\ref{fig:system}, patches from the same row (points in Fig.~\ref{fig:system}a with the same size) or the same column (points in Fig.~\ref{fig:system}b with the same color) exhibit similar attention patterns. Figs.~\ref{fig:system}c and~\ref{fig:system}d reveal further clusters that share similar attention patterns to those in Figs.~\ref{fig:system}a and~\ref{fig:system}b, respectively.

For both modes, we provide three popular dimensionality reduction techniques: PCA, tSNE, and UMAP. Each technique has its unique strengths and limitations, so offering multiple options allows users to explore different perspectives and uncover subtle clustering patterns. When processing very high-dimensional data using tSNE and UMAP, we first use PCA to project them to a relatively lower dimension (i.e., 100D), then apply tSNE and UMAP for efficiency. 
The \textit{Scatterplot} view also supports zooming, enabling users to examine cluster details at various levels of granularity. In Mode 2, users can perform lasso selection to highlight a group of points. The corresponding patches for the selected points will then be visualized in the \textit{Patch} view (described in the following section).

\subsection{The \textit{Patch} View}
The \textit{Patch} view also offers two visualization modes, allowing users to explore patch-level details from the selected stage.

\paragraph{Mode 1:} In this mode, the patches from the selected stage are displayed as gray squares. For example, in Fig.~\ref{fig:system} (right), the $28 {\times} 28$ patches from stage 1 are visualized as $28 {\times} 28$ squares. Meanwhile, the selected patches from the \textit{Scatterplot} view (Fig.~\ref{fig:system}, left) are highlighted as red squares in the \textit{Patch} view, visually indicating their spatial location within the image. This highlighting directly corresponds to the cluster patterns observed in the \textit{Scatterplot} view (Fig.~\ref{fig:system}, left), which helps reveal the spatial relationships between patches that exhibit similar attention patterns. This coordinated visualization provides a clear understanding of how attention patterns are distributed across different regions of the image.

\begin{figure}[tbh]
 \centering 
\includegraphics[width=\columnwidth]{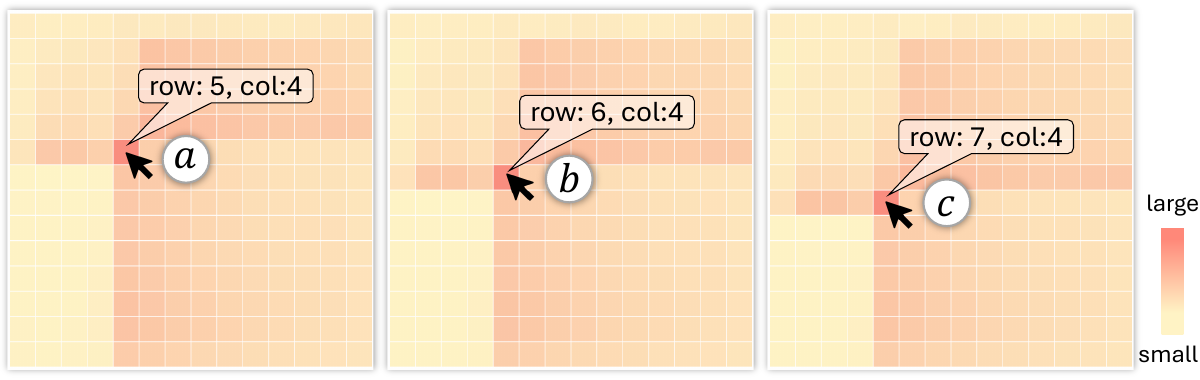}
 \caption{Patches in the same column exhibit similar attention.}
 \label{fig:select_patch}
\end{figure}

\paragraph{Mode 2:} When a square/patch is clicked in the \textit{Patch} view, all squares/patches will be colored according to their attention strength to the clicked one. Attention strengths from small to large are mapped to colors from light-yellow to dark-red. For example, in Fig.~\ref{fig:select_patch}a, the clicked patch is located at row 5, column 4. This patch shows strong attention to: (1) itself, (2) the patches in the same row to its left, and (3) the patches to its right. From the \textit{Scatterplot} view, we noticed that patches in the same column as the selected patch have similar attention patterns. Based on this, we select two additional patches for further exploration. As shown in Figs.~\ref{fig:select_patch}b and~\ref{fig:select_patch}c, these two patches, which are in the same column as the one in Fig.~\ref{fig:select_patch}a, exhibit very similar attention patterns.

\section{Findings and Pattern Summary}
\label{sec:finding}
Using the developed visual analytics tool, we conducted a detailed exploration of the four stages and their respective blocks in the VMamba model (Fig.~\ref{fig:mamba}). This section summarizes the key findings from our analysis.

\begin{figure*}
 \centering 
\includegraphics[width=\textwidth]{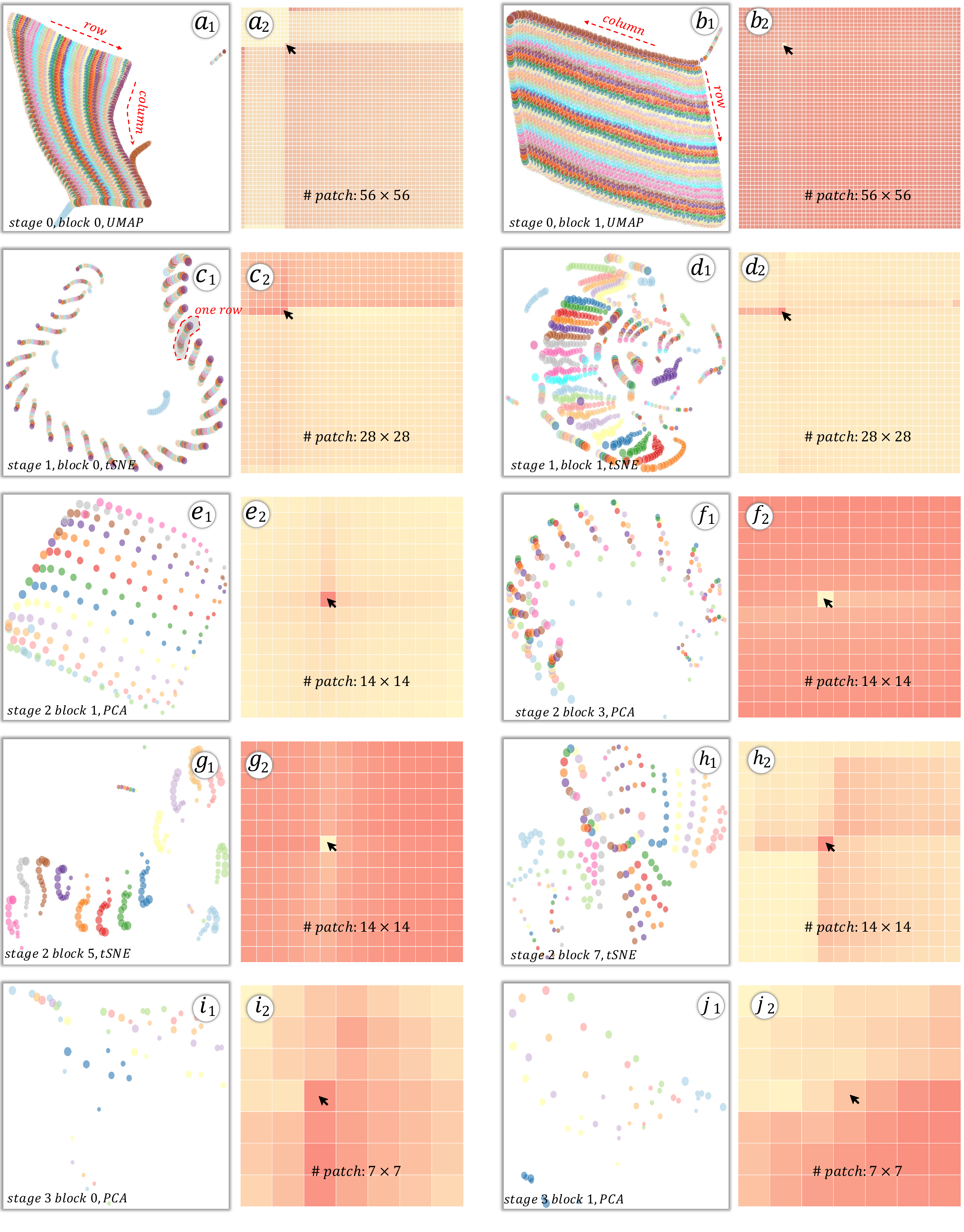}
 \caption{The attention pattern similarity between patches from blocks of different stages. (a1-j1) Two, two, four, and two blocks from stages 0, 1, 2, and 3 are shown, respectively. (a2-j2) Selecting a patch to inspect its attention pattern.}
 \label{fig:pattern}
\end{figure*}

\paragraph{Finding 1:} \textit{Blocks within the same stage exhibit significantly different attention patterns.} This finding is clearly illustrated in Fig.~\ref{fig:stage}. 
To examine the differences further, we focus on the two blocks at stage 0 for a detailed exploration.
As shown in Figs.~\ref{fig:pattern}-a1 (stage 0, block 0) and b1 (stage 0, block 1), the color and size of the points (representing individual patches) change progressively along both the row and column directions. This indicates a gradual transition of the attention pattern for patches along rows and columns.
When individual patches are clicked in the \textit{Patch} view, their attention patterns are displayed on the right. It is evident that the same patch from the two blocks exhibits very different attention patterns. In block 0 (Fig.~\ref{fig:pattern}-a2), the selected patch strongly attends to (1) itself, (2) patches in the first row with a larger column ID, (3) patches in the first column with a larger row ID, and (4) patches that have both larger row and column IDs than the selected patch. In contrast, the patch in block 1 (Fig.~\ref{fig:pattern}-b2) pays noticeably weaker attention to itself and patches along similar rows and columns, but stronger attention to patches in other areas of the image. We checked multiple patches from these two blocks, and the observation is consistent across them.

\paragraph{Finding 2:} \textit{Some blocks exhibit complementary attention patterns.} For instance, the selected patch at stage 2, block 1 (Fig.~\ref{fig:pattern}-e2) shows stronger attention to itself as well as to patches in the same row and column. In contrast, the same patch at stage 2, block 3 (Fig.~\ref{fig:pattern}-f2) and block 5 (Fig.~\ref{fig:pattern}-g2) exhibits weaker attention to the patch itself and to patches along the same row and column. This complementary attention behavior across blocks enables VMamba to selectively focus on relevant patches, contributing to its flexibility in attending to different regions of the input images.

\paragraph{Finding 3:} \textit{There is a hierarchy regarding the attention pattern learned from early to later stages.} At early stages, patches that are spatially closer tend to exhibit similar attention patterns, while at later stages, the attention is more influenced by the image content. At stage 0 (Fig.~\ref{fig:pattern}, a1-b1), attention patterns show smooth and progressive changes for patches in the same row and column, indicating a strong spatial correlation.
At stage 1 (Fig.~\ref{fig:pattern}, c1-d1), clear clusters form for patches in the same row and column—represented by points of the same color but varying sizes or the same size but different colors. 
This pattern persists at stage 2 (Fig.~\ref{fig:pattern}, e1-h1), though clustering is less pronounced for some blocks at this stage. 
By stage 3, the clustering structure becomes less obvious in the \textit{Scatterplot} view (Fig.~\ref{fig:pattern}, i1-j1), suggesting that the attention patterns have become more diverse and content-dependent. 
This phenomenon aligns with the behavior observed in CNNs and vision transformers, where lower layers tend to focus on local, content-agnostic features, while higher layers capture more complex, content-relevant patterns.

\section{Impact of Patch Order on Attention Patterns}
\begin{figure}[tb]
 \centering 
\includegraphics[width=\columnwidth]{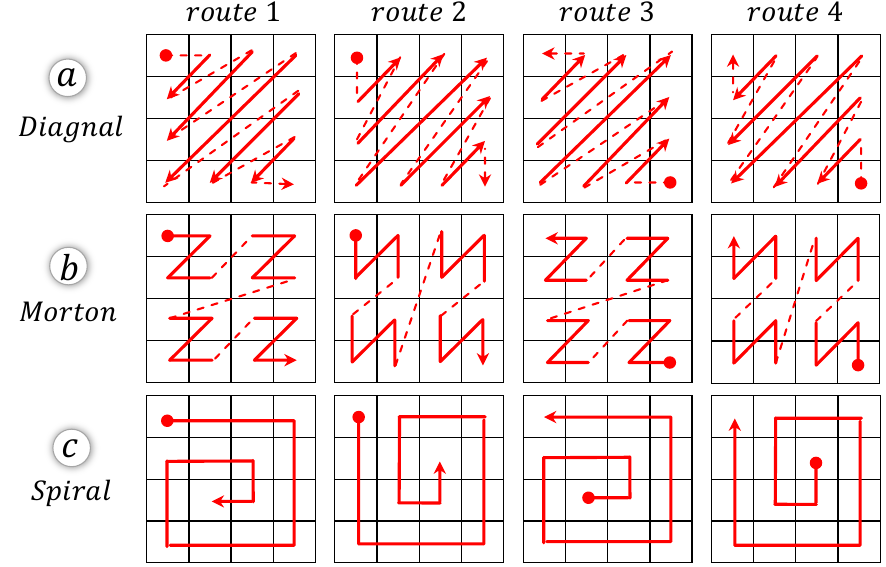}
 \caption{Arranging patches in VMamba following different orders.}
 \label{fig:scan}
\end{figure}

One of our key findings is that patches within the same row or column often exhibit similar attention patterns. We hypothesize that this is influenced by the order in which patches are arranged into sequences. 
To test this hypothesis, we introduce three alternative patch orders, shown in Fig.~\ref{fig:scan}, and examine how the attention patterns change when these new orders are applied.

Fig.~\ref{fig:scan}a shows a patch order that scans patches along the diagonal. Fig.~\ref{fig:scan}b employs the Morton order (z-order curve), a well-known space-filling curve that is particularly effective at preserving spatial locality. Fig.~\ref{fig:scan}c arranges patches in a spiral layout, where the innermost patch retains the highest spatial locality.
We modified the VMamba code to implement each of these patch orders and trained the model from scratch for 400 epochs. All three scanning methods achieved accuracy levels similar to the original VMamba model, i.e., achieving accuracy greater than or equal to 82.6\% on ImageNet as reported in the original VMamba paper~\cite{liu2024vmamba}.

\begin{figure}[tbh]
 \centering 
\includegraphics[width=\columnwidth]{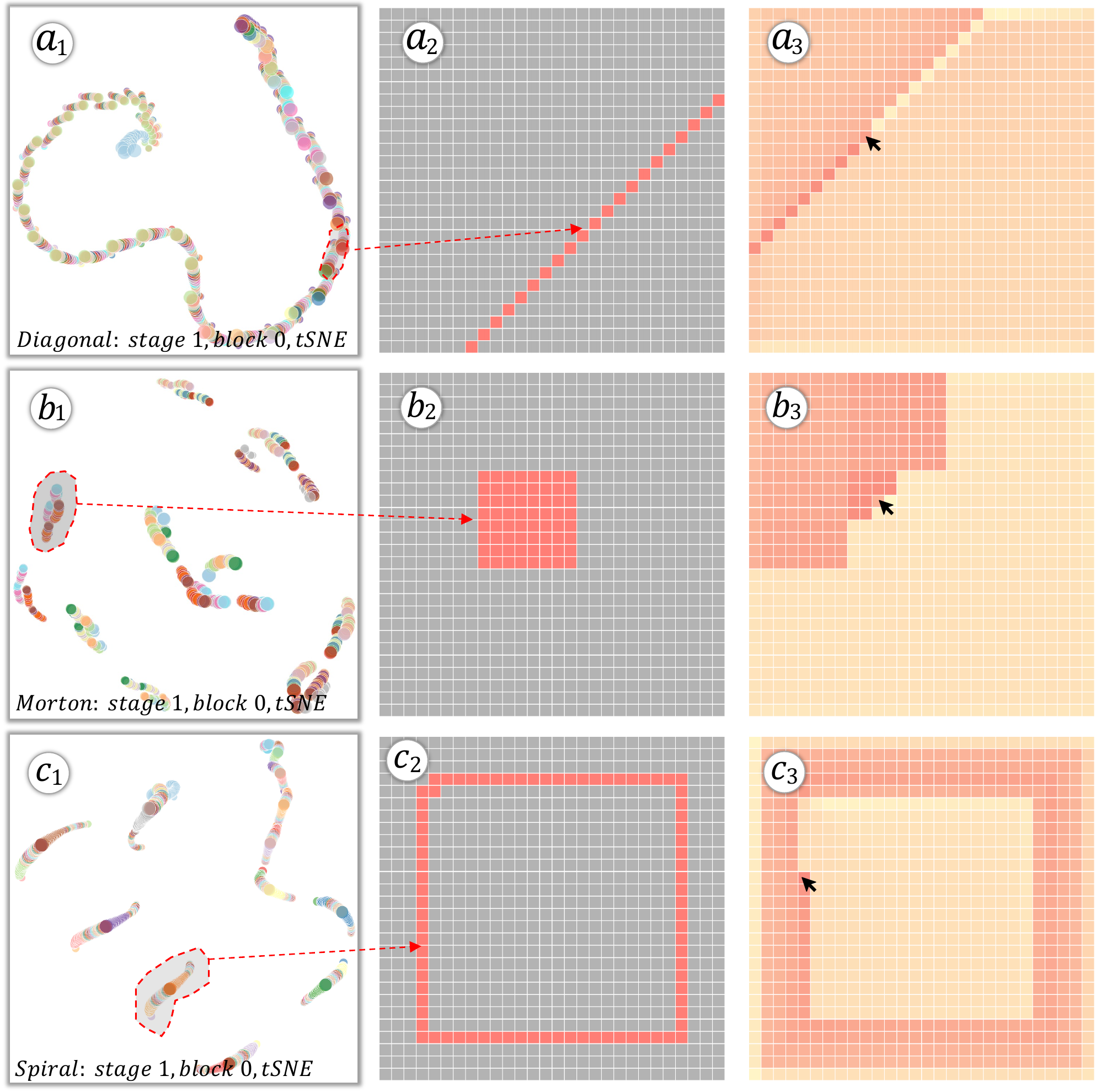}
 \caption{Exploring attention patterns when patches are arranged following the three new orders in Fig.~\ref{fig:scan}: Diagonal, Morton, and Spiral.}
 \label{fig:pattern_more}
\end{figure}

Figs.~\ref{fig:pattern_more}-a1, a2, and a3 show the results of exploring attention patterns using the diagonal order. In Fig.~\ref{fig:pattern_more}-a1, all patches are projected onto a continuous curve. By selecting points in a local region, we observe that the corresponding patches are adjacent along the diagonal, as shown in Fig.~\ref{fig:pattern_more}-a2. When clicking on any patch in the \textit{Patch} view, we see that its attention behavior mirrors what was observed in Fig.~\ref{fig:pattern} but along the diagonals instead of the rows or columns. For example, the attention pattern of a patch at stage 1, block 0 is displayed in Fig.~\ref{fig:pattern_more}-a3. This patch strongly attends to its preceding patches along the diagonal, which is very similar to what was observed in Fig.~\ref{fig:pattern}-c2. However, in Fig.~\ref{fig:pattern}-c2, the preceding patches are those based on the cross-scan order in Fig.~\ref{fig:patch}. 

Next, we conduct similar explorations with the VMamba model trained using the Morton order. As shown in Fig.~\ref{fig:pattern_more}-b1, patches at stage 1, block 0 are grouped into clusters. Selecting a cluster highlights the patches within a local region in Fig.~\ref{fig:pattern_more}-b2. These patches are actually in the same z-order curve block. The strong spatial locality among them results in similar attention patterns. Clicking on a patch reveals that it also strongly attends to its preceding patches, as shown in Fig.~\ref{fig:pattern_more}-b3. 
The results for the spiral patch order, shown in Figs.~\ref{fig:pattern_more}-c1, c2, and c3, are consistent with those observed from other patch orders.
These findings confirm that stage 1, block 0 of the VMamba model exhibits a fixed attention pattern: any patch in this block consistently attends strongly to its preceding patches. However, the definition of ``preceding patches" depends on the specific patch order used.

\section{Conclusion and Future Work}

In this paper, we introduced a visual analytics tool to explore and compare attention patterns within and across VMamba blocks. Through our exploration, we discovered several key insights: (1) VMamba blocks within the same stage exhibit distinct attention patterns; (2) the order of patches significantly influences the resulting attention patterns; and (3) patches that are close in the input sequence generally exhibit similar attention patterns.
Given the significant impact of patch arrangement on attention patterns, we proposed multiple new patch orders that better preserve the patches' spatial locality. Using our tool, we further investigated VMamba models trained with these new orders, and found similar attention behaviors to those observed in the original patch order.

Looking ahead, we aim to extend our tool in several directions. First, we plan to explore more complex VMamba models with additional stages and blocks per stage. We hypothesize that the general trends observed in this study will hold as the architecture becomes more intricate. Second, our current work focuses on identifying content-agnostic attention patterns by averaging attention over $n$ images. In the future, we intend to integrate additional views into the tool to present content-relevant attention patterns, which will be particularly useful for diagnosing specific images of interest.


\bibliographystyle{abbrv-doi}

\bibliography{template}
\end{document}